\documentclass[runningheads]{llncs}

\usepackage{graphicx}

\usepackage{amsmath}

\usepackage{amsfonts}
\usepackage{bbold}

\usepackage{amsthm}
\usepackage{amssymb}
\usepackage{url}
\usepackage{pifont}

\usepackage{subcaption}
\begin{document}
\title{Master your Metrics with Calibration}
%
%
\author{Wissam Siblini\orcidID{0000-0002-4193-2061} \and
Jordan Fréry \and
Liyun He-Guelton \and
Frédéric Oblé \and
Yi-Qing Wang}
\authorrunning{W. Siblini et al.}
%
\institute{Worldline, France\\ \email{firstname.lastname@worldline.com}}
\maketitle              
\begin{abstract}
Machine learning models deployed in real-world applications are often evaluated with precision-based metrics such as F1-score or AUC-PR (Area Under the Curve of Precision Recall). Heavily dependent on the class prior, such metrics make it difficult to interpret the variation of a model's performance over different subpopulations/subperiods in a dataset. In this paper, we propose a way to calibrate the metrics so that they can be made invariant to the prior. We conduct a large number of experiments on balanced and imbalanced data to assess the behavior of calibrated metrics and show that they improve interpretability and provide a better control over what is really measured. We describe specific real-world use-cases where calibration is beneficial such as, for instance, model monitoring in production, reporting, or fairness evaluation.

\keywords{Performance Metrics  \and Class Imbalance \and Precision-Recall}
\end{abstract}
\section{Introduction}

\begin{figure*}[ht]
\centering
\includegraphics[width=0.70\textwidth]{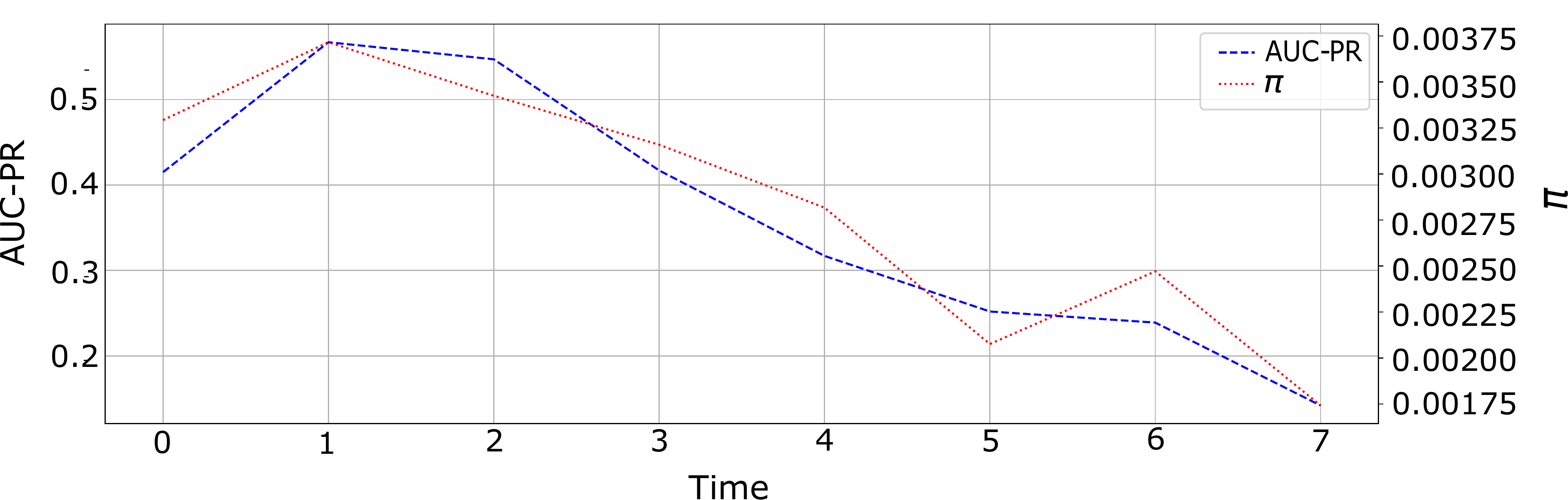}

\caption{Evolution of the AUC-PR of a fraud detection system and of the fraud ratio ($\pi$, i.e. the empirical $\mathbb{P}_t(y)$)) over time. Both decrease, but, as the AUC-PR is dependent on the prior, it does not allow to tell if the performance variation is only due to the variation of $\pi$ or if there was a drift in $\mathbb{P}_t(x|y)$}\label{fraud_ratio_perd}
\end{figure*}

In real-world machine learning systems, the predictive performance of a model is often evaluated on multiple datasets, and comparisons are made. These datasets can correspond to sub-populations in the data, or different periods in time~\cite{tatbul2018precision}. Choosing the best suited metrics is not a trivial task. Some metrics may prevent a proper interpretation of the performance differences between the sets~\cite{garcia2012suitability,santafe2015dealing}, especially because different datasets generally not only have a different likelihood $\mathbb{P}(x|y)$ but also a different class prior $\mathbb{P}(y)$. A metric dependent on the prior (e.g. precision) will be affected by both differences indiscernibly~\cite{brzezinski2019dynamics} but a practitioner could be interested in isolating the variation of performance due to likelihood which reflects the intrinsic model's performance (see illustration in Figure \ref{fraud_ratio_perd}). Take the example of comparing the performance of a model across time periods: At time $t$, we receive data drawn from $\mathbb{P}_t(x, y) = \mathbb{P}_t(x|y) \mathbb{P}_t(y)$ where $x$ are the features and $y$ the label. Hence the optimal scoring function (i.e. model) for this dataset is the likelihood ratio \cite{neyman1933ix}:
\begin{align}\label{scoring_function}
s_t(x) := \frac{\mathbb{P}_t(x|y=1)}{\mathbb{P}_t(x|y=0)}
\end{align}
In particular, if $\mathbb{P}_t(x|y)$ does not vary with time, neither will $s_t(x)$. In this case, even if the prior $\mathbb{P}_t(y)$ varies, it is desirable to have a performance metric $M(\cdot)$ satisfying $M(s_t, \mathbb{P}_t) = M(s_{t+1}, \mathbb{P}_{t+1}), \forall t$ so that the model maintains the same metric value over time. That being said, this does not mean that dependence to prior is an intrinsically bad behavior. Some applications seek this property as it reflects a part of the difficulty to classify on a given dataset (e.g. the performance of the random classifier evaluated with a prior-dependent metric is more or less high depending on the skew of the dataset).

In binary classification, researchers often rely on the AUC-ROC (Area Under the Curve of Receiver Operating Characteristic) to measure a classifier's performance~\cite{hanley1982meaning,fawcett2006introduction}. While this metric has the advantage of being invariant to the class prior, many real-world applications, especially when data are imbalanced, have recently begun to favor precision-based metrics such as AUC-PR and F-Score\cite{saito2015precision}\cite{sajjadi2018assessing}. The reason is that AUC-ROC suffers from giving false positives too little importance~\cite{davis2006relationship} although the latter strongly deteriorate user experience and waste human efforts with false alerts. Indeed AUC-ROC considers a tradeoff between TPR and FPR whereas AUC-PR/F1-score consider a tradeoff between TPR (Recall) and Precision. With a closer look, the difference boils down to the fact that it normalizes the number of false positives with respect to the number of true negatives whereas precision-based metrics normalize it with respect to the number of true positives. In highly imbalanced scenarios (e.g. fraud/disease detection), the first is much more likely than the second because negative examples are in large majority. 

Precision-based metrics give false positives more importance, but they are tied to the class prior~\cite{branco2016survey,brzezinski2019dynamics}. A new definition of precision and recall into precision gain and recall gain has been recently proposed to correct several drawbacks of AUC-PR \cite{flach2015precision}. But, while the resulting AUC-PR Gain has some advantages of the AUC-ROC such as the validity of linear interpolation between points, it remains dependent on the class prior. Our study aims at providing metrics (i) that are precision-based to tackle problems where the class of interest is highly under-represented and (ii) that can be made independent of the prior for comparison purposes (e.g. monitoring the evolution of the performance of a classifier accross several time periods). To reach this objective, this paper provides: (1) A formulation of calibration for precision-based metrics. It compute the value of precision as if the ratio $\pi$ of the test set was equal to a reference class ratio $\pi_0$. We give theoretical arguments to explain why it allows invariance to the class prior. We also provide a calibrated version for precision gain and recall gain \cite{flach2015precision}. (2) An empirical analysis on both synthetic and real-world data to confirm our claims and show that new metrics are still able to assess the model's performance and are easier to interpret. (3) A large scale experiments on 614 datasets using openML \cite{vanschoren2014openml} to (a) give more insights on correlations between popular metrics by analyzing how they rank models, (b) explore the links between the calibrated metrics and the regular ones.

Not only calibration solves the issue of dependence to the prior but also allows, with parameter $\pi_0$, anticipating a different ratio and controlling what the metric precisely reflects. This new property has several practical interests (e.g. for development, reporting, analysis) and we discuss them in realistic use-cases in section \ref{guideline}.

\section{Popular Metrics for Binary Classification: Advantages and Limits}\label{sec:related_met}

We consider a usual binary classification setting where a model has been trained and its performance is evaluated on a test dataset of $N$ instances. $y_i \in \{0,1\}$ is the ground-truth label of the $i^{\text{th}}$ instance and is equal to $1$ (resp. $0$) if the instance belongs to the positive (resp. negative) class. The model provides $s_i \in \mathbb{R}$, a score for the $i^{\text{th}}$ instance to belong to the positive class. For a given threshold $\tau \in \mathbb{R}$, the predicted label is $\widehat{y}_i = 1$ if $s_i > \tau$ and $0$ otherwise. Predictive performance is generally measured using the number of true positives ($\text{TP} =\sum_{i=1}^N \mathbb{1}(\widehat{y}_i = 1, y_i = 1)$), true negatives ($\text{TN}  = \sum_{i=1}^N \mathbb{1}(\widehat{y}_i = 0, y_i = 0)$), false positives ($\text{FP} =  \sum_{i=1}^N \mathbb{1}(\widehat{y}_i = 1, y_i = 0)$), false negatives ($\text{FN}  =\sum_{i=1}^N \mathbb{1}(\widehat{y}_i = 0, y_i = 1)$). One can compute relevant ratios such as the True Positive Rate (TPR) also referred to as the Recall ($Rec = \frac{\text{TP}}{\text{TP}+\text{FN}}$), the False Positive Rate ($\text{FPR} = \frac{\text{FP}}{\text{TN}+\text{FP}}$) also referred to as the Fall-out and the Precision ($Prec = \frac{\text{TP}}{\text{TP}+\text{FP}}$). As these ratios are biased towards a specific type of error and can easily be manipulated with the threshold, more complex metrics have been proposed. In this paper, we discuss the most popular ones which have been widely adopted in binary classification: F1-Score, AUC-ROC, AUC-PR and AUC-PR Gain. F1-Score is the harmonic average between $Prec$ and $Rec$:

\begin{equation}
\label{eq_f1}
F_1 = \frac{2*Prec*Rec}{Prec+Rec}.
\end{equation}

\begin{figure*}[htb]

    \centering 
    $\pi = 0.003$ \\
\begin{subfigure}{0.31\textwidth}
  \includegraphics[width=\linewidth]{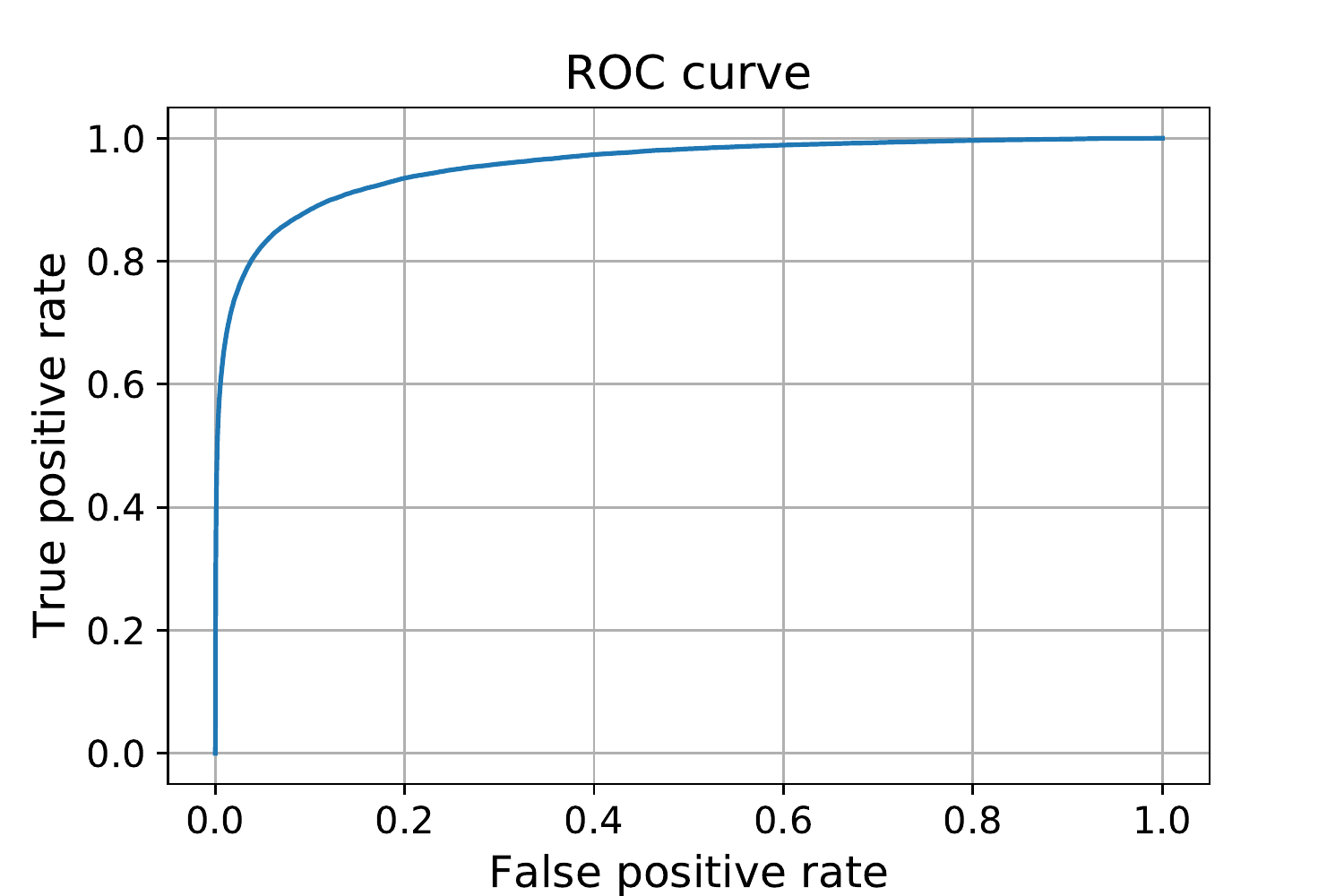}

\end{subfigure}\hfil 
\begin{subfigure}{0.31\textwidth}
  \includegraphics[width=\linewidth]{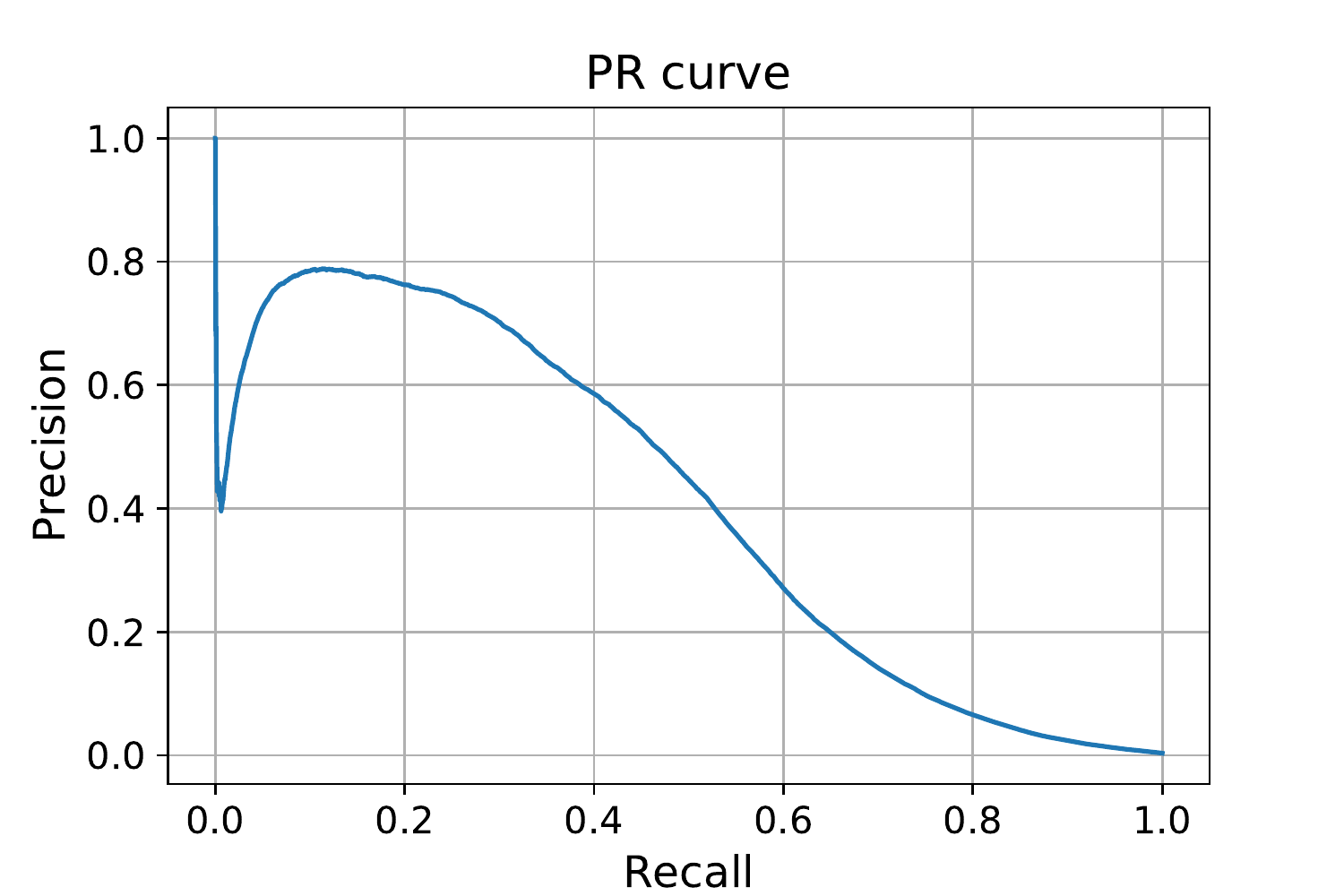}

\end{subfigure}\hfil 
\begin{subfigure}{0.31\textwidth}
  \includegraphics[width=\linewidth]{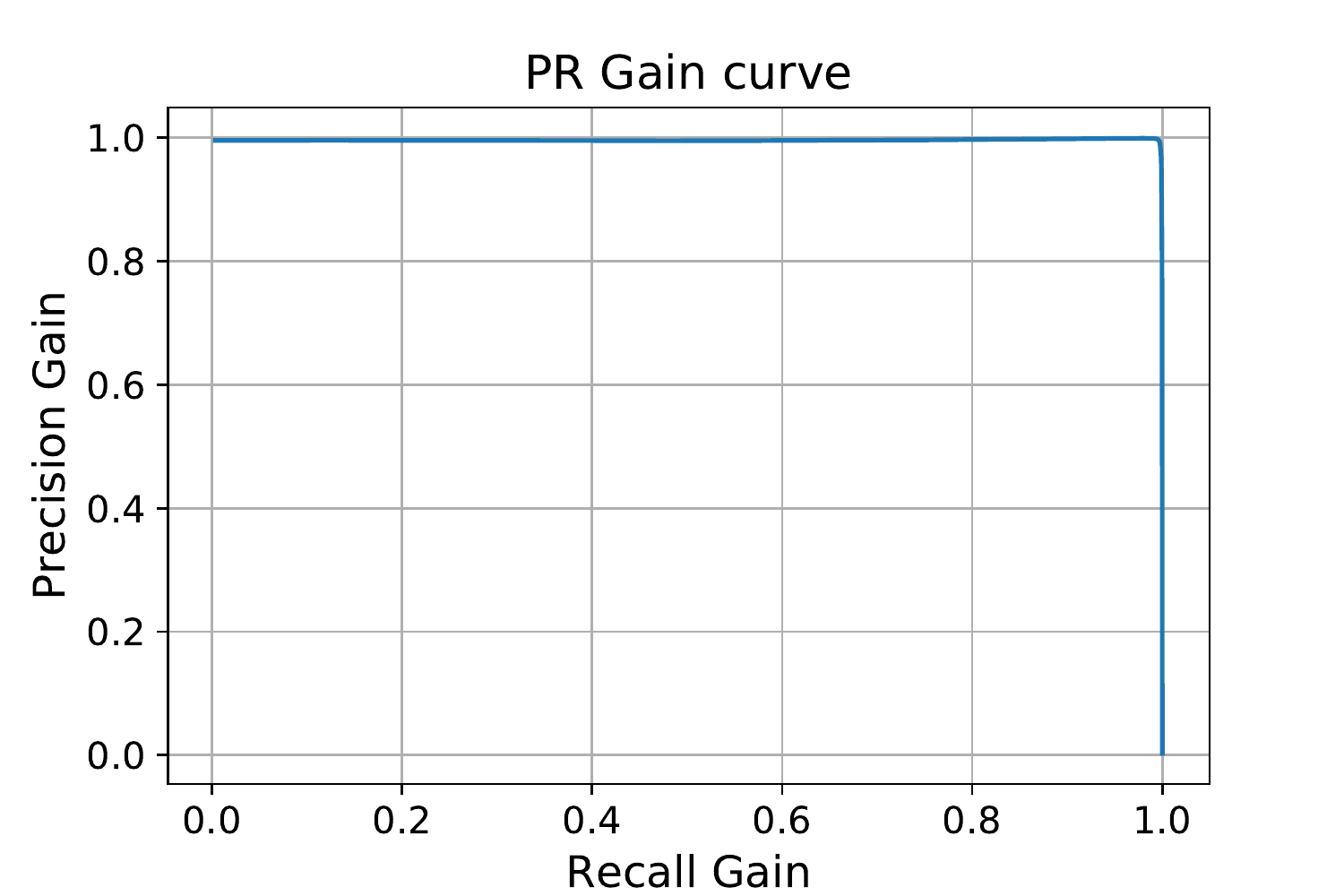}

\end{subfigure}

 $\pi = 0.5$ \\
\begin{subfigure}{0.31\textwidth}
  \includegraphics[width=\linewidth]{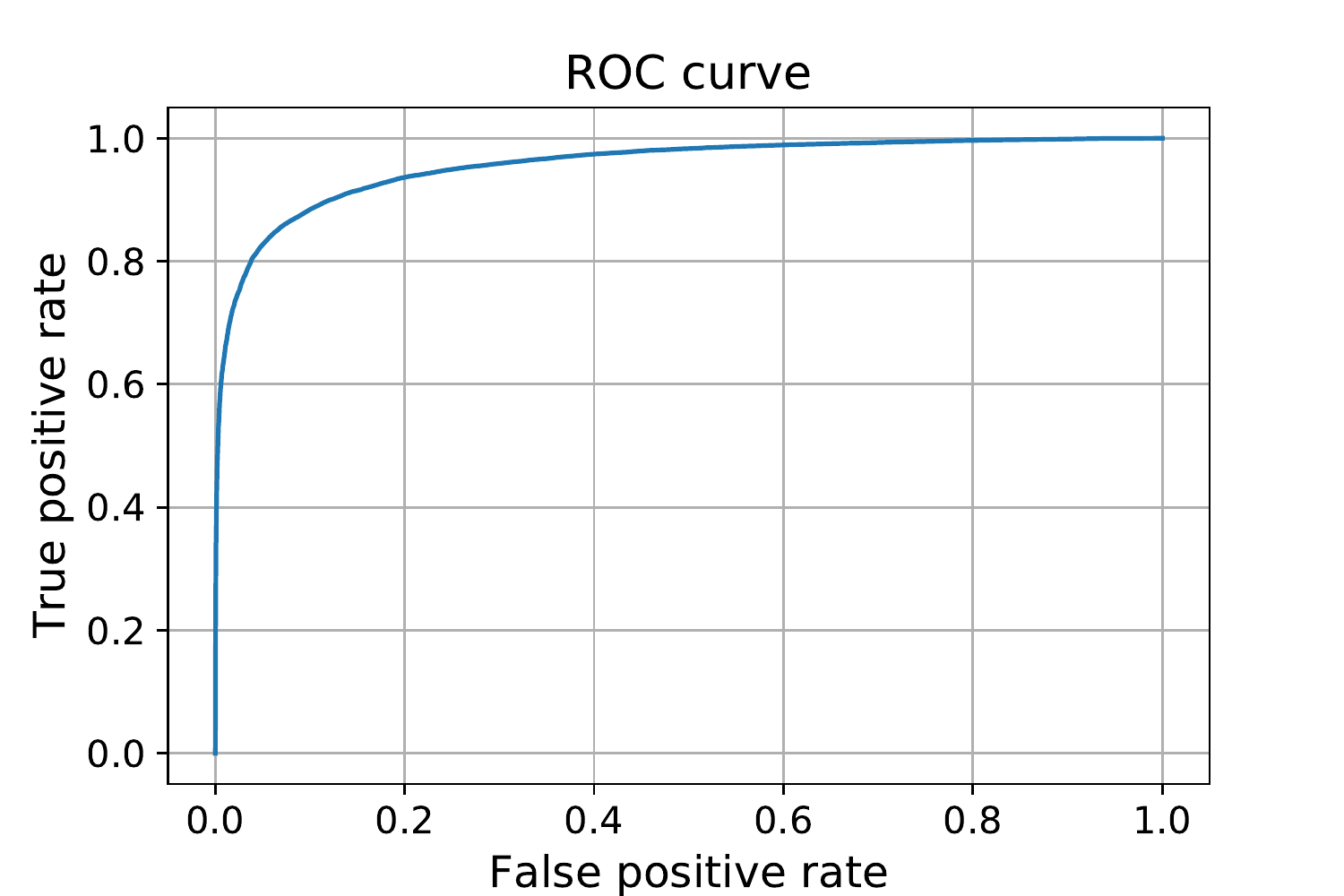}

\end{subfigure}\hfil 
\begin{subfigure}{0.31\textwidth}
  \includegraphics[width=\linewidth]{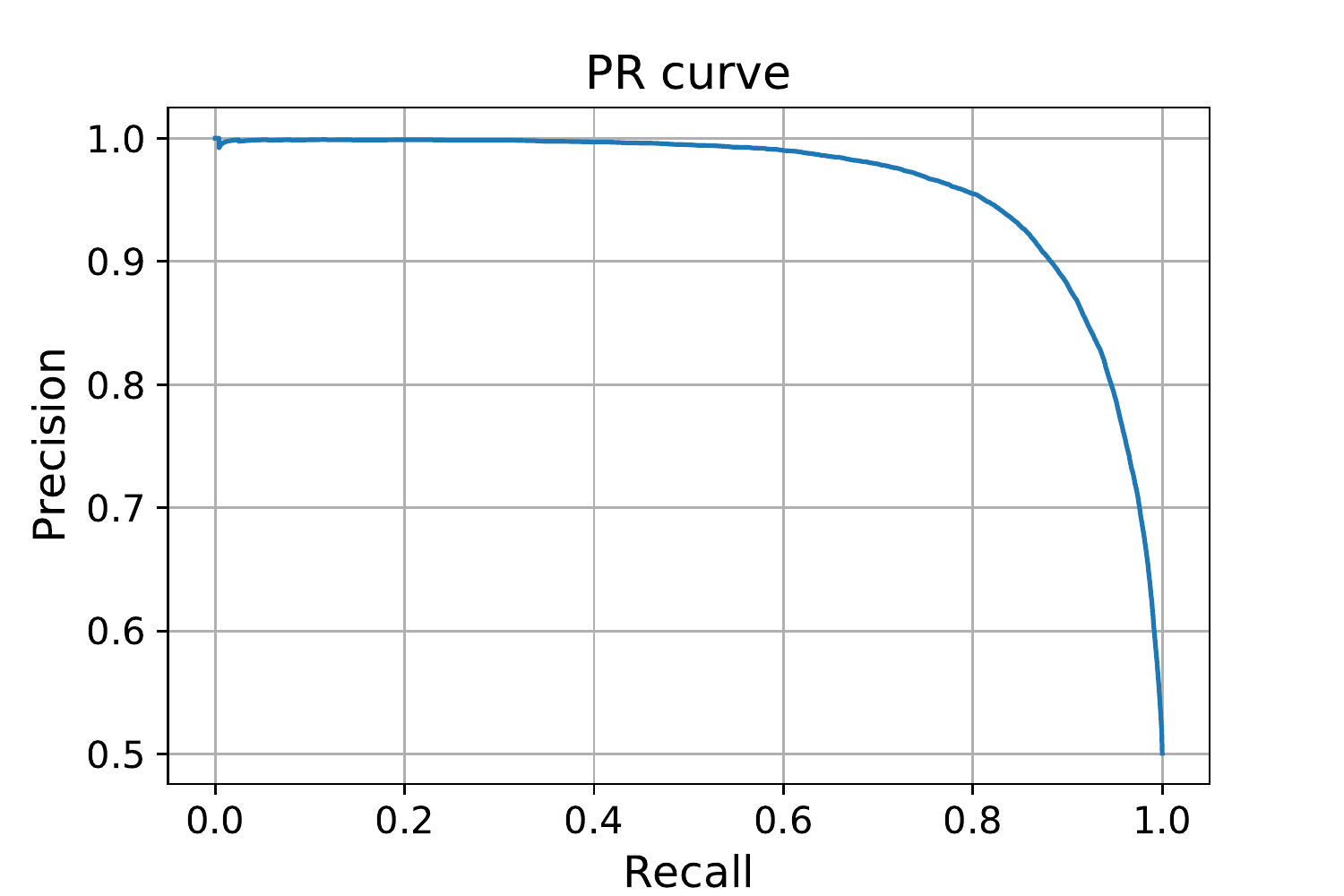}

\end{subfigure}\hfil 
\begin{subfigure}{0.31\textwidth}
  \includegraphics[width=\linewidth]{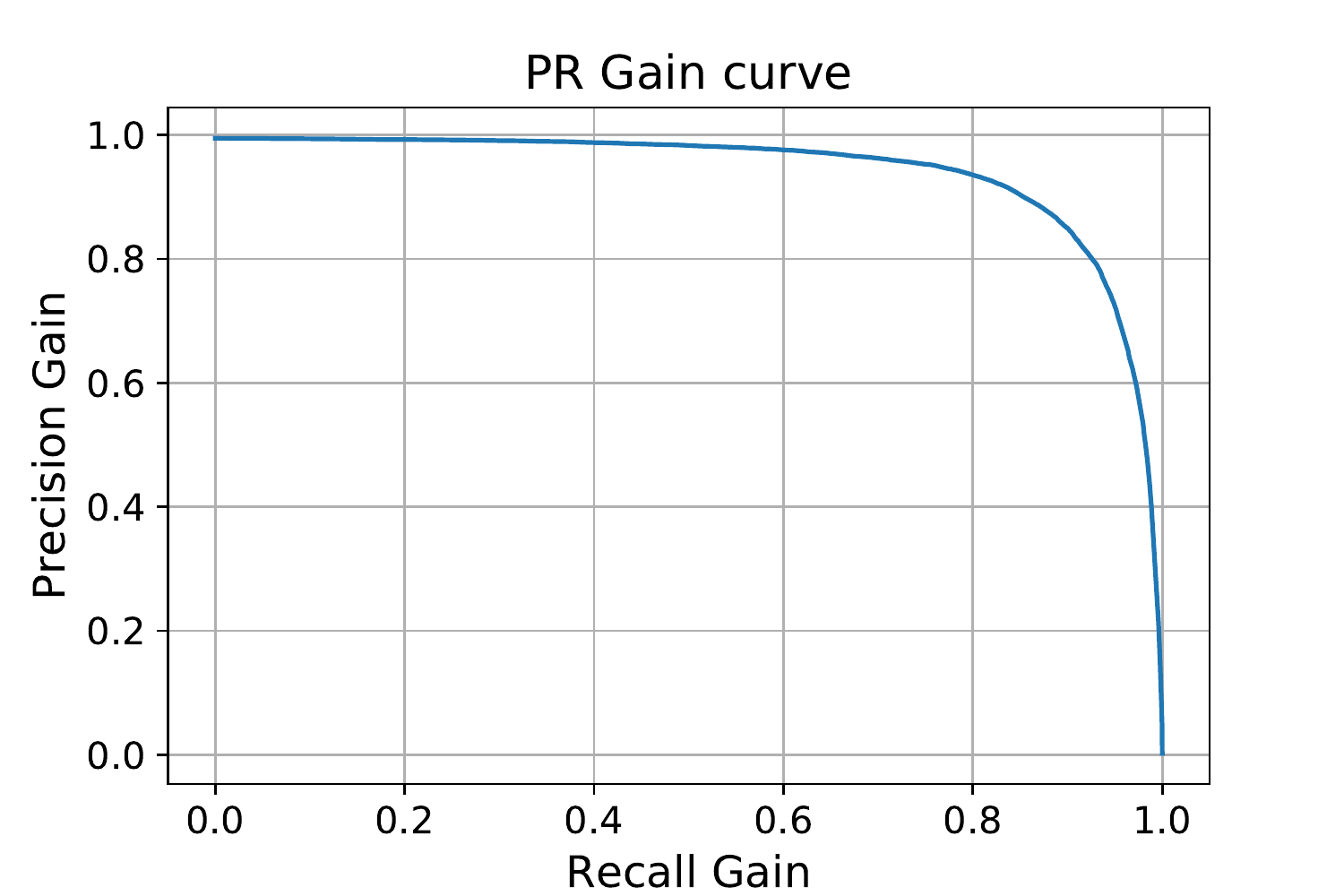}

\end{subfigure}
\caption{ROC, PR and PR gain curves for the same model evaluated on an extremely imbalanced test set from a fraud detection application ($\pi = 0.003$, in the top row) and on a balanced sample ($\pi = 0.5$, in the bottom row).}
\label{fig:images}
\end{figure*}

The three other metrics consider every threshold $\tau$ from the highest $s_i$ to the lowest. For each one, they compute TP, FP, TN and FN. Then, they plot one ratio against another and compute the Area Under the Curve (Figure \ref{fig:images}). AUC-ROC considers the Receiver Operating Characteristic curve where TPR is plotted against FPR. AUC-PR considers the Precision vs Recall curve. Finally, in AUC-PR Gain, the precision gain ($Prec_G$) is plotted against the recall gain ($Rec_G$). They are defined in \cite{flach2015precision} as follows ($\pi = \frac{\sum_{i=1}^N y_i}{N}$ is the positive class ratio and we always consider that it is the minority class in this paper): 

\begin{minipage}{0.45\columnwidth}
\begin{equation}
\label{eq_prec_gain}
Prec_G = \frac{Prec - \pi}{(1-\pi)Prec}
\end{equation}
\end{minipage}%
\begin{minipage}{0.45\columnwidth}
\begin{equation}
\label{eq_rec_gain}
Rec_G = \frac{Rec - \pi}{(1-\pi)Rec}
\end{equation}
\end{minipage}

PR Gain enjoys many properties of the ROC that the regular PR analysis does not (e.g. the validity of linear interpolations or the existence of universal baselines) \cite{flach2015precision}. However, AUC-PR Gain becomes hardly usable in extremely imbalanced settings. In particular, we can derive from \eqref{eq_prec_gain} and \eqref{eq_rec_gain} that $Prec_G$/$Rec_G$ will be mostly close to 1 if $\pi$ is close to 0 (see top right chart in Figure \ref{fig:images}).

\begin{figure*}[ht]

\centering
\includegraphics[width=0.85\textwidth]{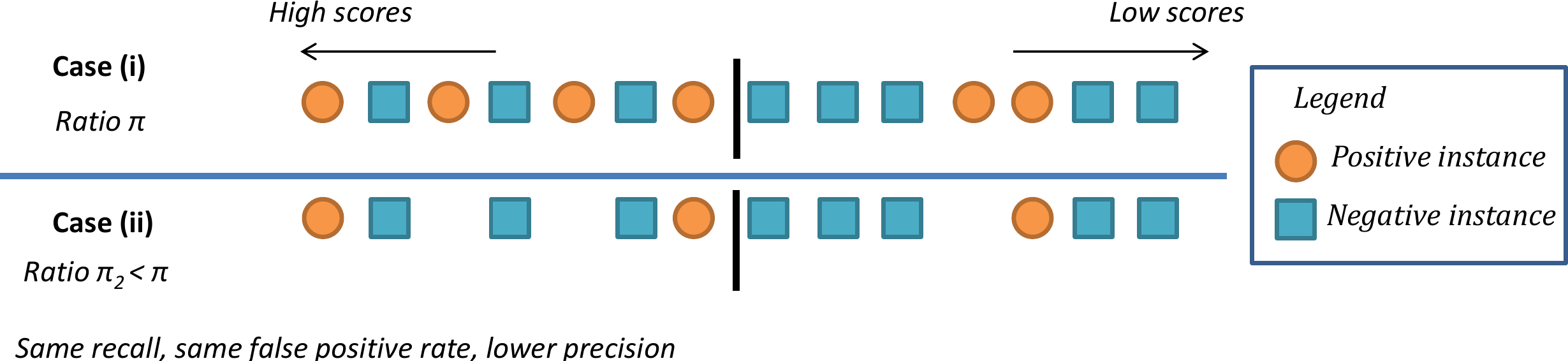}
\caption{Illustration of the impact of $\pi$ on precision, recall, and the false positive rate. Instances are ordered from left to right according to their score given by the model. The threshold is illustrated as a vertical line between the instances: those on the left (resp. right) are classified as positive (resp. negative)}
\label{schema_pr_pourcent}

\end{figure*}

As explained in the introduction, precision-based metrics (F1, AUC-PR) are more adapted than AUC-ROC for problems with class imbalance. On the other hand, only AUC-ROC is invariant to the positive class ratio. Indeed, FPR and $Rec$ are both unrelated to the class ratio because they only focus on one class but it is not the case for $Prec$. Its dependency on the positive class ratio $\pi$ is illustrated in Figure \ref{schema_pr_pourcent}: when comparing a case (i) with a given ratio $\pi$ and another case (ii) where a randomly selected half of the positive examples has been removed, one can visually understand that both recall and false positive rate are the same but the precision is lower in the second case.

\section{Calibrated Metrics}

We seek a metric that is based on $Prec$ to tackle problems where data are imbalanced and the minority (positive) class is the one of interest but we want it to be invariant w.r.t the class prior to be able to interpret its variation across different datasets (e.g. different time periods). To obtain such a metric, we will modify those based on $Prec$ (AUC-PR, F1-Score and AUC-PR Gain) to make them independent of the positive class ratio $\pi$.

\subsection{Calibration}
\label{subseccalib}
 The idea is to fix a reference ratio $\pi_0$ and to weigh the count of TP or FP in order to calibrate them to the value that they would have if $\pi$ was equal to $\pi_0$. $\pi_0$ can be chosen arbitrarily (e.g. $0.5$ for balanced) but it is preferable to fix it according to the task at hand (we analyze the impact of $\pi_0$ in section \ref{interpretation_metrics} and describe simple guidelines to fix it in section \ref{guideline}). 

If the positive class ratio is $\pi_0$ instead of $\pi$, the ratio between negative examples and positive examples is multiplied by $\frac{\pi(1-\pi_0)}{\pi_0(1-\pi)}$. In this case, we expect the ratio between false positives and true positives to be multiplied by $\frac{\pi(1-\pi_0)}{\pi_0(1-\pi)}$. Therefore, we define the calibrated precision $Prec_c$ as follows: 

\begin{equation}
\label{calib_metric}
Prec_c = \frac{\text{TP}}{\text{TP} + \frac{\pi(1-\pi_0)}{\pi_0(1-\pi)}\text{FP}} = \frac{1}{1 + \frac{\pi(1-\pi_0)}{\pi_0(1-\pi)}\frac{\text{FP}}{\text{TP}}}
\end{equation}

Since $\frac{1-\pi}{\pi}$ is the imbalance ratio $\frac{N_-}{N_+}$ where $N_+$ (resp. $N_-$) is the number of positive (resp. negative) examples, we have: $\frac{\pi}{1-\pi}\frac{\text{FP}}{\text{TP}} = \frac{\text{FP}/N_-}{\text{TP}/N_+} = \frac{\text{FPR}}{\text{TPR}}$ which is independent of $\pi$. 

Based on the calibrated precision, we can also define the calibrated F1-score, the calibrated $Prec_G$ and the calibrated $Rec_G$ by replacing $Prec$ by $Prec_c$ and $\pi$ by $\pi_0$ in equations \eqref{eq_f1}, \eqref{eq_prec_gain} and \eqref{eq_rec_gain}. Note that calibration does not change precision gain. Indeed, calibrated precision gain $\frac{Prec_c - \pi_0}{(1-\pi_0)Prec_c}$ can be rewritten as $\frac{Prec - \pi}{(1-\pi)Prec}$ which is equal to the regular precision gain. Also, the interesting properties of the recall gain were proved independently of the ratio $\pi$ in \cite{flach2015precision} which means that calibration preserves them.

\subsection{Robustness to Variations in $\pi$}
\label{subsecrobus}

In order to evaluate the robustness of the new metrics to variations in $\pi$, we create a synthetic dataset where the label is drawn from a Bernoulli distribution with parameter $\pi$ and the feature is drawn from Normal distributions: 
\begin{equation}\label{synthetic_datatset}
\begin{aligned}
p(x|y=1;\mu_1) = \mathcal{N}(x;\mu_1,1), \hspace{1cm}
p(x|y=0;\mu_0) = \mathcal{N}(x;\mu_0,1)
\end{aligned}
\end{equation}

\begin{figure*}[h]
\centering
\includegraphics[width=0.90\textwidth]{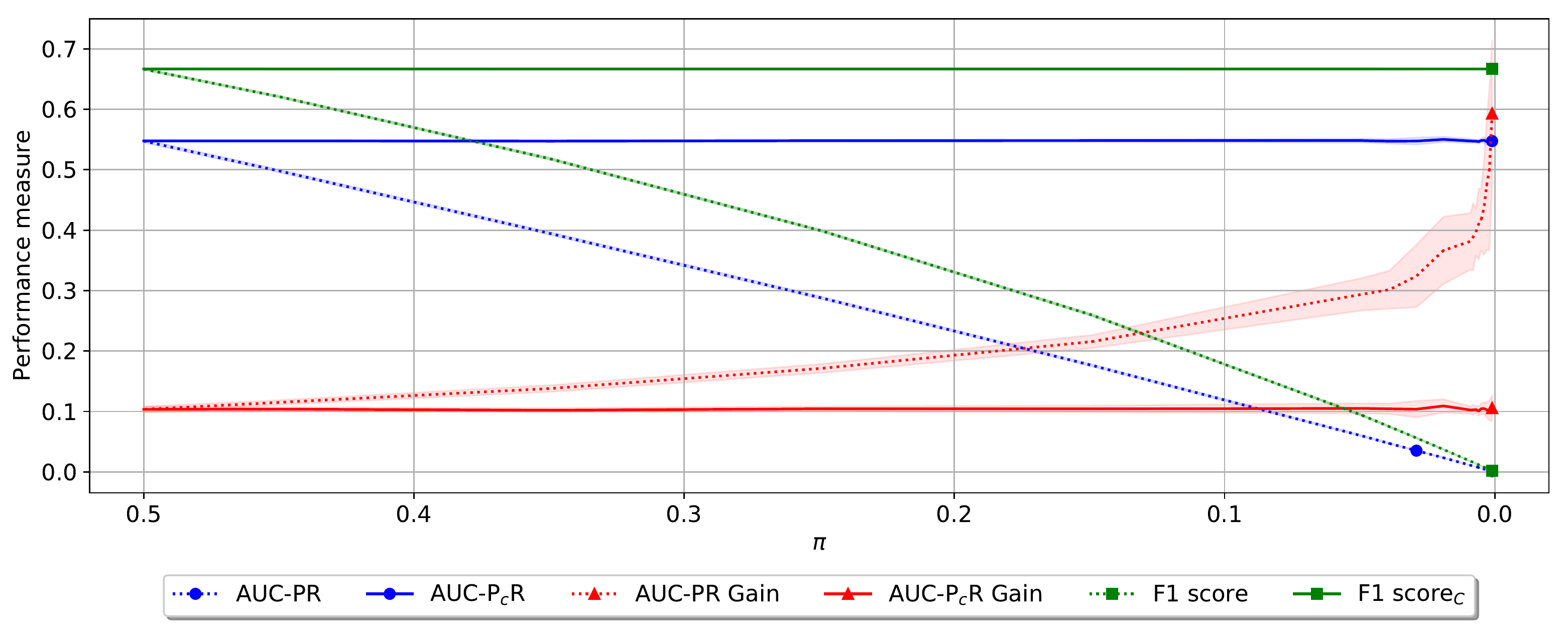}
\caption{Evolution of AUC-PR, AUC-PR Gain, F1-score and their calibrated version (AUC-P$_c$R, AUC-P$_c$R Gain, F1-score$_c$) as $\pi$ decreases. We arbitrarily set $\pi_0 = 0.5$ for the calibrated metrics. The curves are obtained by averaging results over 30 runs and we show the confidence intervals.}

\label{evo_metric_ratio_corr}

\end{figure*}

For several values of $\pi$, data points are generated from \eqref{synthetic_datatset} with $\mu_1 = 2$ and $\mu_0 = 1.8$. We consider a large number of points ($10^6$) so that the empirical class ratio $\pi$ is approximately equal to the Bernouilli parameter $\pi$. We empirically study the evolution of several metrics ($F_1$-score, AUC-PR, AUC-PR Gain and their calibrated version) for the optimal model (as defined in \eqref{scoring_function}) as $\pi$ decreases from $\pi = 0.5$ (balanced) to $\pi = 0.001$. We observe that the impact of the class prior on the regular metrics is important (Figure~\ref{evo_metric_ratio_corr}). It can be a serious issue for applications where $\pi$ sometimes vary by one order of magnitude from one day to another (see~\cite{dal2018credit} for a real world example) as it leads to a significant variation of the measured performance (see the difference between AUC-PR when $\pi = 0.5$ and when $\pi=0.05$) even if the optimal model remains the same. On the contrary, the calibrated versions remain very robust to changes in the class prior $\pi$ even for extreme values. Note that we here experiment with synthetic data to have a full control over the distribution/prior and make the analysis easier but the conclusions are exactly the same on real world data.\footnote{see appendix in \url{https://figshare.com/articles/Calibrated_metrics_IDA_Supplementary_material_pdf/11848146}\label{footnote_see_appendix}}

\subsection{Assessment of the Model Quality}
\label{subsecassess}

Besides the robustness of the calibrated metrics to changes in $\pi$, we also want them to be sensitive to the quality of the model. If this latter decreases regardless of the $\pi$ value, we expect all metrics, calibrated ones included, to decrease in value. Let us consider an experiment where we use the same synthetic dataset as defined the previous section. However, instead of changing the value of $\pi$ only, we change $(\mu_1,\mu_0)$ to make the problem harder and harder and thus worsen the optimal model's performance. This can be done by reducing the distance between the two normal distributions in \eqref{synthetic_datatset}, because this would result in more overlapping between the classes and make it harder to discriminate between them. As a distance, we consider the \textit{KL}-divergence that boils down to $\frac{1}{2}(\mu_1 - \mu_0)^2$. 

 \begin{figure*}[h]
 \centering
\includegraphics[width=0.90\textwidth]{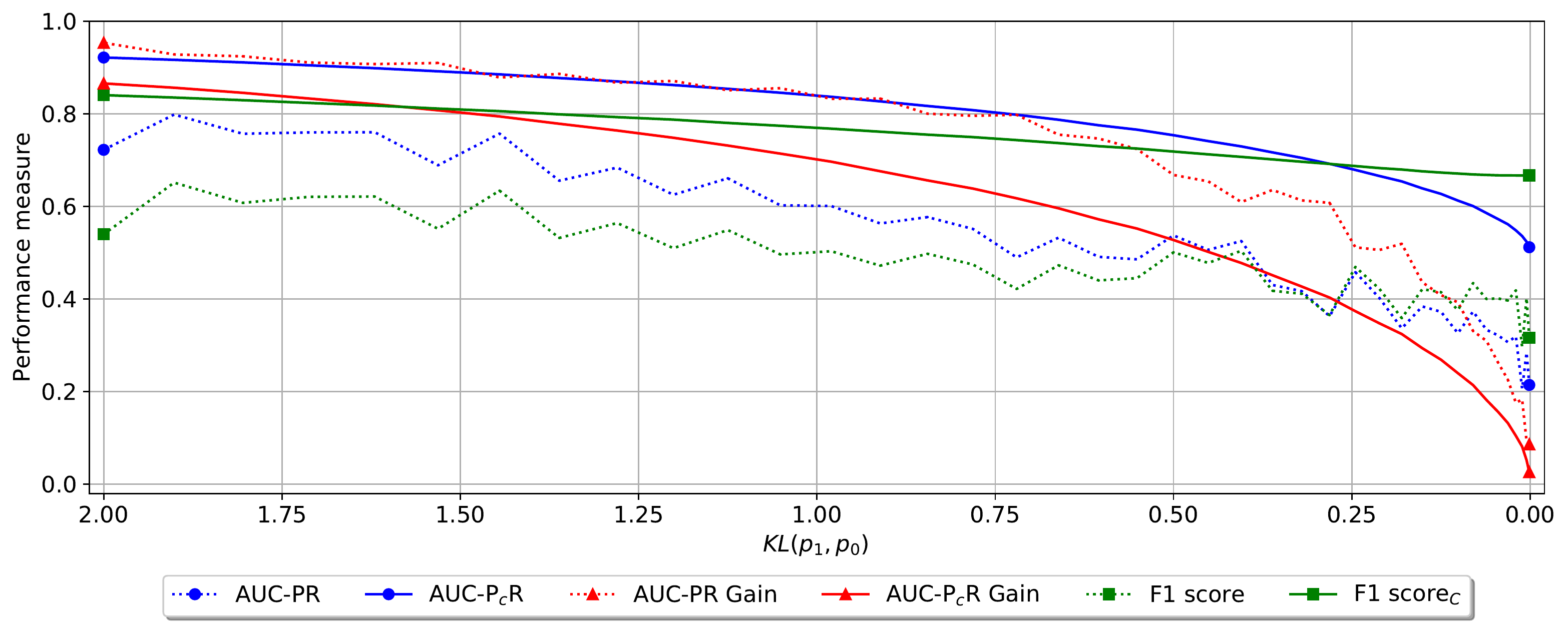}
\caption{Evolution of AUC-PR, AUC-PR Gain, F1-score and their calibrated version as $KL(p_1,p_0)$ tends to 0 and as $\pi$ randomly varies. This curve was obtained by averaging results over 30 runs.}\label{worse_models}
\end{figure*}

Figure~\ref{worse_models} shows how the values of the metrics evolve as the KL-divergence gets closer to zero. For each run, we randomly chose the prior $\pi$ in the interval $[0.001,0.5]$. As expected, all metrics globally decrease as the problem gets harder. However, we can notice an important difference: the variation in the calibrated metrics are smooth and monotonic compared to those of the original metrics which are affected by the random changes in $\pi$. In that sense, variations of the calibrated metrics across the different generated datasets are much easier to interpret than the original metrics.

\section{Link Between Calibrated and Original Metrics}\label{interpretation_metrics}

\subsection{Meaning of $\pi_0$}

Let us first remark that for test datasets in which $\pi = \pi_0$, $Prec_c$ is equal to the regular precision $Prec$ since $\frac{\pi(1-\pi_0)}{\pi_0(1-\pi)}=1$ (this is observable in Figure~\ref{evo_metric_ratio_corr} with the intersection of the metrics for $\pi = \pi_0 = 0.5$). 

\begin{figure}[h]

\centering
\includegraphics[width=0.5\textwidth]{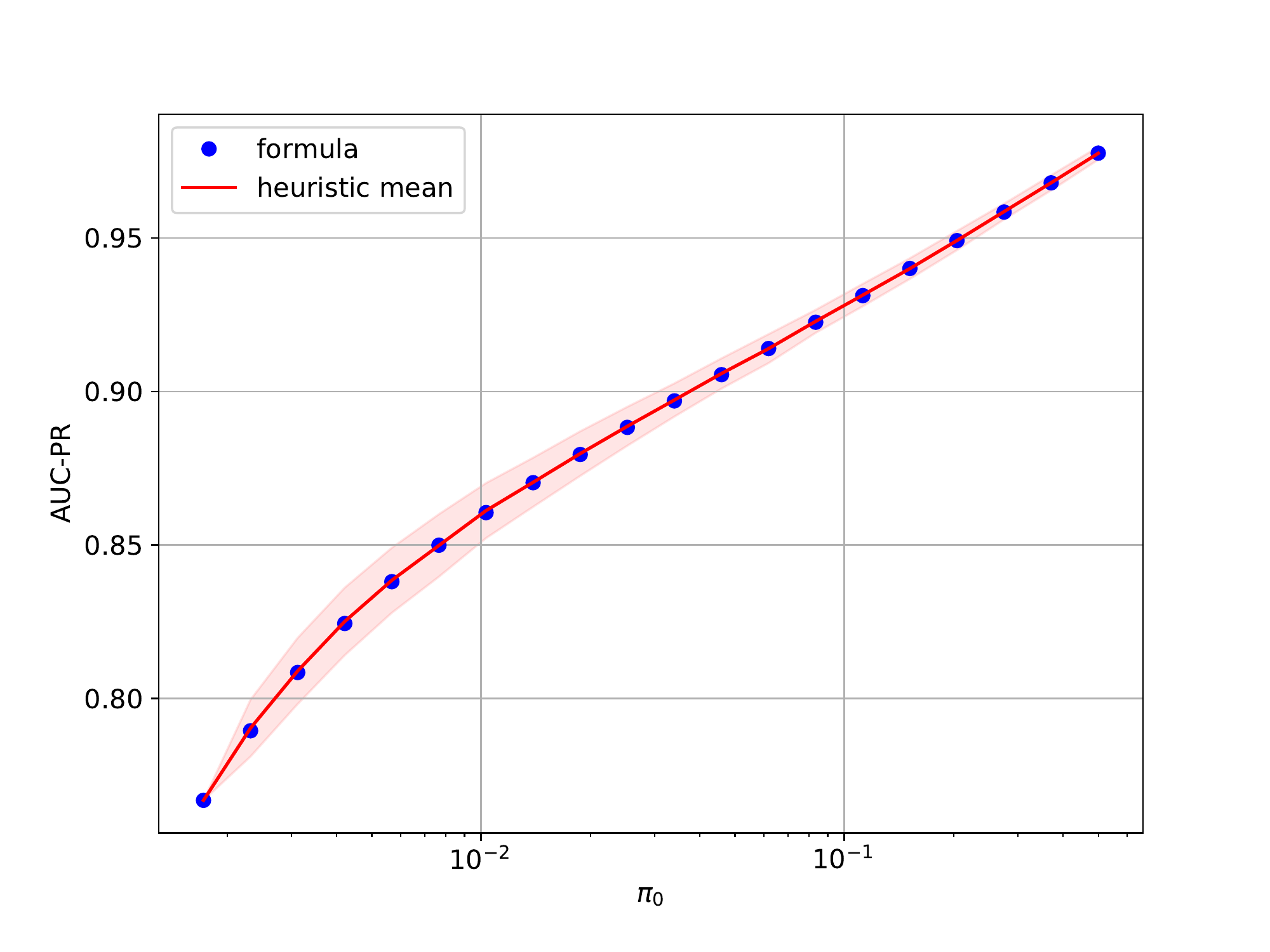}
\caption{Comparison between heuristic-based calibrated AUC-PR (red line) and our closed-form calibrated AUC-PR (blue dots). The red shadow represents the standard deviation of the heuristic-based calibrated AUC-PR over 1000 runs.}\label{heuristic_vs_formula}
\end{figure}

If $\pi \neq \pi_0$, the calibrated metrics essentially have the value that the original ones would have if the positive class ratio $\pi$ was equal to $\pi_0$. To further demonstrate that, we compare our proposal for calibration \eqref{calib_metric} with the only proposal from the past \cite{jeni2013facing} that was designed for the same objective: a heuristic-based calibration. The approach from \cite{jeni2013facing} consists in randomly undersampling the test set to make the positive class ratio $\pi$ equal to a chosen ratio (let us refer to it as $\pi_0$ for the analogy) and then computing the regular metrics on the sampled set. Because of the randomness, sampling may remove more hard examples than easy examples so the performance can be over-estimated, and vice versa. To avoid that, the approach performs several runs and computes a mean estimation. In Figure \ref{heuristic_vs_formula}, we compare the results obtained with our formula and with their heuristic, for several reference ratio $\pi_0$, on a highly unbalanced ($\pi = 0.0017$) credit card fraud detection dataset available on Kaggle \cite{dal2018credit}. 

We can observe that our formula and the heuristic provide really close values. This can be theoretically explained (see appendix - footnote \ref{footnote_see_appendix}) and confirms that the calibration formula computes the value that the original metric would have if the ratio $\pi$ in the test set was $\pi_0$. Note that our closed-form calibration \eqref{calib_metric} can be seen as an improvement of the heuristic-based calibration from \cite{jeni2013facing} as it provides the targeted value without running a costly Monte-Carlo simulation. 

\subsection{Do the Calibrated Metrics Rank Models in the Same Order as the Original Metrics ?}

\begin{figure*}[h]
\centering
\includegraphics[width=\textwidth]{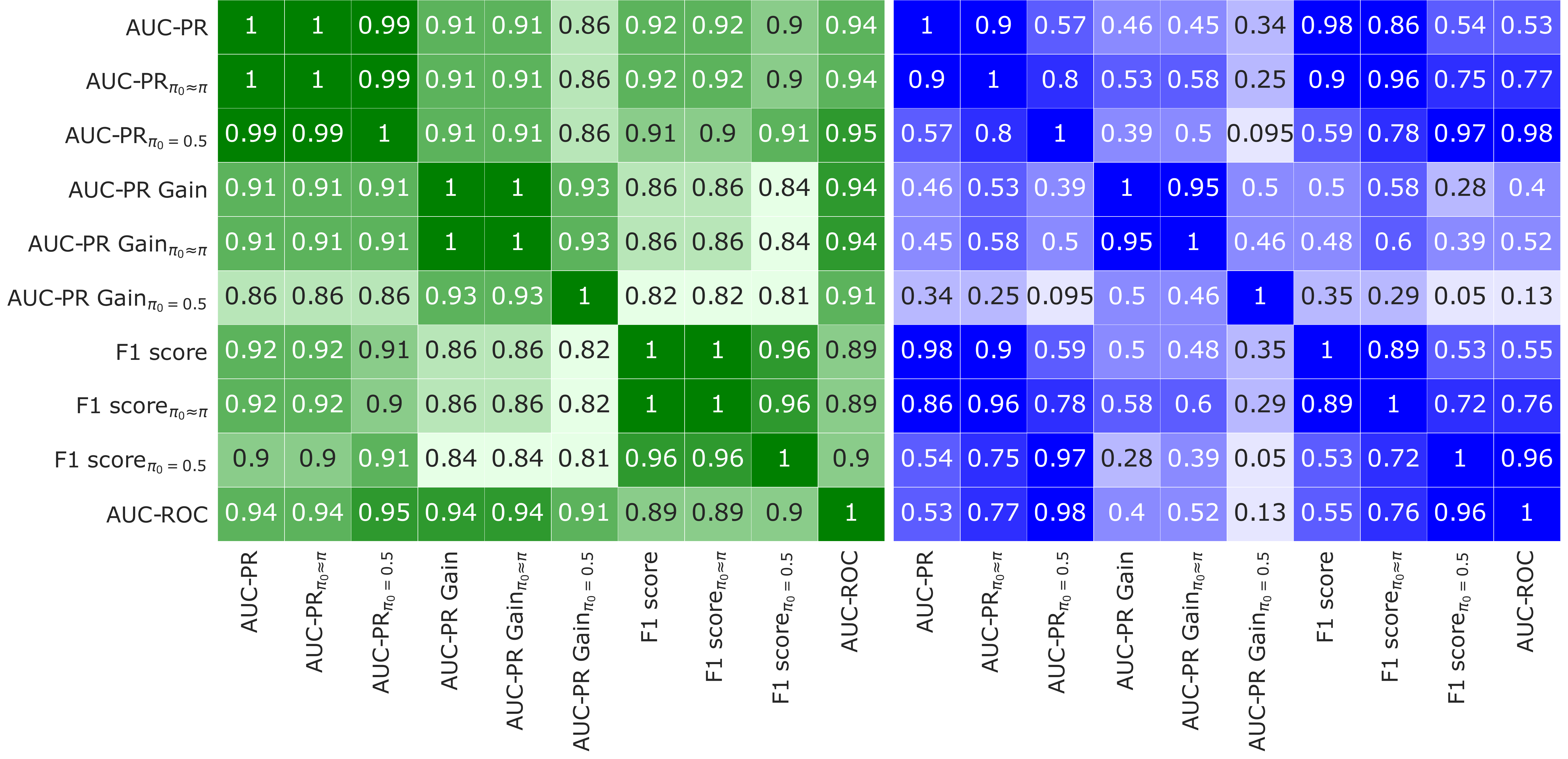}
\caption{Spearman rank correlation matrices between 10 metrics over 614 datasets for the left figure and the 4 highly imbalanced datasets for the right figure.}\label{correlation_matrices}
\end{figure*}

Calibration results in evaluating the metric for a different prior. In this section, we analyze how this impacts the task of selectioning the best model for a given dataset. To do this, we empirically analyze the correlation of several metrics in terms of model ordering. We use OpenML \cite{vanschoren2014openml} to select the 602 supervised binary classification datasets on which at least 30 models have been evaluated with a 10-fold cross-validation. For each one, we randomly choose 30 models, fetch their predictions, and evaluate their performance with the metrics. This leaves us with $614\times30=18,420$ different values for each metric. To analyze whether they rank the models in the same order, we compute the Spearman rank correlation coefficient between them for the 30 models for each of the $614$ problems.\footnote{the implementation of the paper experiments can be found at \url{https://github.com/wissam-sib/calibrated_metrics}} Most datasets roughly have balanced classes ($\pi > 0.2$ in more than 90\% of the datasets). Therefore, to also specifically analyze the imbalance case, we run the same experiment with only the subset of $4$ highly imbalanced datasets ($\pi < 0.01$). The compared metrics are AUC-ROC, AUC-PR, AUC-PR Gain and the best F1-score over all possible thresholds. We also add the calibrated version of the last three. In order to understand the impact of $\pi_0$, we use two different values: the arbitrary $\pi_0 = 0.5$ and another value $\pi_0 \approx \pi$ (for the first experiment with all datasets, $\pi_0 \approx \pi$ corresponds to $\pi_0 = 1.01\pi$ and for the second experiment where $\pi$ is very small, we go further and $\pi_0 \approx \pi$ corresponds to $\pi_0 = 10\pi$ which remains closer to $\pi$ than $0.5$). The obtained correlation matrices are shown in Figure~\ref{correlation_matrices}. Each individual cell corresponds to the average Spearman correlation over all datasets between the row metric and the column metric.

A general observation is that most metrics are less correlated with each other when classes are unbalanced (right matrix in Figure~\ref{correlation_matrices}). We also note that the best F1-score is more correlated to AUC-PR than to AUC-ROC or AUC-PR Gain. In the balanced case (left matrix in Figure~\ref{correlation_matrices}), we can see that metrics defined as area under curves are generally more correlated with each other than with the threshold sensitive classification metric F1-score. Let us now analyze the impact of calibration. As expected, in general, when $\pi_0 \approx \pi$, calibrated metrics have a behavior really close to that of the original metrics because $\frac{\pi(1-\pi_0)}{\pi_0(1-\pi)}\approx 1$ and therefore $Prec_c \approx Prec$. In the balanced case (left), since $\pi$ is close to $0.5$, calibrated metrics with $\pi_0 = 0.5$ are also highly correlated with the original metrics. In the imbalanced case (on the right matrix of Figure~\ref{correlation_matrices}), when $\pi_0$ is arbitrarily set to $0.5$ the calibrated metrics seem to have a low correlation with the original ones. In fact, they are less correlated with them than with AUC-ROC. And this makes sense given the relative weights that each of the metric applies to FP and TP. The original precision gives the same weight to $TP$ and $FP$, although false positives are $\frac{1-\pi}{\pi}$ times more likely to occur ($\frac{1-\pi}{\pi} > 100$ if $\pi < 0.01$). The calibrated precision with the arbitrary value $\pi_0 = 0.5$ boils down to $\frac{\text{TP}}{\text{TP} + \frac{\pi}{(1-\pi)}\text{FP}}$ and gives a weight $\frac{1-\pi}{\pi}$ times smaller to false positives which counterbalances their higher likelihood. ROC, like the calibrated metrics with $\pi_0 = 0.5$, gives $\frac{1-\pi}{\pi}$ less weight to FP because it is computed from FPR and TPR which are linked to $\text{TP}$ and $\text{FP}$ with the relationship $\frac{\pi}{1-\pi}\frac{\text{FP}}{\text{TP}} = \frac{\text{FPR}}{\text{TPR}}$.

To sum up the results, we first emphasize that the choice of the metrics to rank classifiers when datasets are rather balanced seems to be much less sensitive than in the extremely imbalanced case. In the balanced case the least correlated metrics have an average rank correlation of $0.81$. For the imbalanced datasets, on the other hand, many metrics have low correlations which means that they often disagree on the best model. The choice of the metric is therefore very important here. Our experiment also seems to reflect that rank correlations are mainly a matter of how much weight is given to each type of error. Choosing these "weights" generally depends on the application at hand. An this should be remembered when using calibration. To preserve the nature of a given metrics, $\pi_0$ has to be fixed to a value close to $\pi$ and not arbitrarily. The user still has the choice to fix it to another value if his purpose is to specifically place the results into a different reference with a different prior.

\section{Guidelines and Use-Cases}
\label{guideline}

Calibration could benefit ML practitioners when analyzing the performance of a model across different datasets/time periods. Without being exhaustive, we give four use-cases where it is beneficial (setting $\pi_0$ depends on the target use-case):

\textbf{Comparing the performance of a model on two populations/classes:} Consider a practitioner who wants to predict patients with a disease and evaluate the performance of his model on subpopulations of the dataset (e.g. children, adults and elderly people). If the prior is different from one population to another (e.g. elderly people are more likely to have the disease), precision will be affected, i.e. population with a higher disease ratio will be more likely to have a higher precision. In this case, the calibrated precision can be used to obtain the precision of each population set to the same reference prior (for instance, $\pi_0$ can be chosen as the average prior over all populations). This would provide an additional balanced point of view and make the analysis richer to draw more precise conclusions and perhaps study fairness \cite{barocas2017fairness}. 

\textbf{Model performance monitoring in an industrial context:} in systems where a model's performance is monitored over time with precision-based metrics like F1-score, using calibration in addition to the regular metrics makes it easier to understand the evolution especially when the class prior can evolve (cf. application in Figure \ref{fraud_ratio_perd}). For instance, it can be useful to analyze the drift (i.e. distinguish between variations linked to $\pi$ or $P(X|y)$) and design adapted solutions; either updating the threshold or completely retraining the model. To avoid denaturing too much the F1-score, here $\pi_0$ has to be fixed based on realistic values (e.g. average $\pi$ in historical data).

\textbf{Establishing agreements with clients:} As shown in previous sections, $\pi_0$ can be interpreted as the ratio to which we refer to compute the metric. This can be useful to establish a guarantee, in an agreement, that will be robust to uncontrollable events. Indeed, if we take the case of fraud detection, the real positive class ratio $\pi$ can vary extremely from one day to another and on particular events (e.g. fraudster attacks, holidays) which significantly affects the measured metrics (see Figure \ref{evo_metric_ratio_corr}). Here, after having both parties to agree beforehand on a reasonable value for $\pi_0$ (based on their business knowledge), calibration will always compute the performance relative to this ratio and not the real $\pi$ and thus be easier to guarantee. 

\textbf{Anticipating the deployment of a model in production:} Imagine one collects a sample of data to develop an algorithm and reaches an acceptable AUC-PR for production. If the prior in the collected data is different from reality, the non-calibrated metric might have given either a pessimistic or optimistic estimation of the post-deployment performance. This can be extremely harmful if the production has strict constraints. Here, if the practitioner uses calibration with $\pi_0$ equal to the minimal prior envisioned for the application at hand, he/she would be able to 
anticipate the worst case scenario.
\section{Conclusion}

\label{sec:disc}

In this paper, we provided a formula of calibration, empirical results, and guidelines to make the values of metrics across different datasets more interpretable. Calibrated metrics are a generalization of the original ones. They rely on a reference $\pi_0$ and compute the value that we would obtain if the positive class ratio $\pi$ in the evaluated test set was equal to $\pi_0$. If the user chooses $\pi_0 = \pi$, this does not change anything and he retrieves the regular metrics. But, with different choices, the metrics can serve several purposes such as obtaining robustness to variation in the class prior across datasets, or anticipation. They are useful in both academic and industrial applications as explained in the previous section: they help drawing more accurate comparisons between subpopulations, or study incremental learning on streams by providing a point of view agnostic to virtual concept drift \cite{widmer1993effective}. They can be used to provide more controllable performance indicators (easier to guarantee and report), help preparing deployment in production, and prevent false conclusions about the evolution of a deployed model. However, $\pi_0$ has to be chosen with caution as it controls the relative weights given to FP and TP and, consequently, can affect the selection of the best classifier. 

\bibliographystyle{splncs04}
\bibliography{references}

\end{document}